\documentclass{sig-alternate-05-2015}
\usepackage[
   pass, 
]{geometry}

\usepackage{graphicx}
\usepackage{subfigure}
\usepackage{multirow}
\usepackage{color}

\begin{document}

\title{Transform-Invariant Convolutional Neural Networks for Image Classification and Search}
\numberofauthors{5} 
\author{
\alignauthor Xu Shen \\
       \affaddr{CAS Key Laboratory of Technology in Geo-spatial Information Processing and Application System}\\
       \affaddr{University of Science and Technology of China}\\
       \affaddr{Hefei, Anhui, China 230027}\\
       \email{shenxu@mail.ustc.edu.cn}
\alignauthor Xinmei Tian \\
       \affaddr{CAS Key Laboratory of Technology in Geo-spatial Information Processing and Application System}\\
       \affaddr{University of Science and Technology of China}\\
       \affaddr{Hefei, Anhui, China 230027}\\
       \email{xinmei@ustc.edu.cn}
\alignauthor Anfeng He \\
       \affaddr{CAS Key Laboratory of Technology in Geo-spatial Information Processing and Application System}\\
       \affaddr{University of Science and Technology of China}\\
       \affaddr{Hefei, Anhui, China 230027}\\
       \email{heanfeng@mail.ustc.edu.cn}
\and  
\alignauthor Shaoyan Sun\\
       \affaddr{CAS Key Laboratory of Technology in Geo-spatial Information Processing and Application System}\\
       \affaddr{University of Science and Technology of China}\\
       \affaddr{Hefei, Anhui, China 230027}\\
       \email{sunshy@mail.ustc.edu.cn}
\alignauthor Dacheng Tao\\
       \affaddr{ Centre for Quantum Computation \& Intelligent Systems and the Faculty of Engineering and Information Technology}\\
       \affaddr{University of Technology, Sydney}\\
       \affaddr{Ultimo, NSW 2007, Australia}\\
       \email{dacheng.tao@uts.edu.au}
}

\maketitle
\begin{abstract}
Convolutional neural networks (CNNs) have achieved state-of-the-art results on many visual
recognition tasks. However, current CNN models still exhibit a poor ability to be invariant to
spatial transformations of images. Intuitively, with sufficient layers and parameters,
hierarchical combinations of convolution (matrix multiplication and non-linear activation)
and pooling operations should be able to learn a robust mapping from transformed input
images to transform-invariant representations. In this paper, we propose randomly transforming
 (rotation, scale, and translation) feature maps of CNNs during the training stage. This prevents complex
 dependencies of specific rotation, scale, and translation levels of training images in CNN models. Rather,
 each convolutional kernel learns to detect a feature that is generally helpful for producing the
 transform-invariant answer given the combinatorially large variety of transform levels of its input
 feature maps. In this way, we do not require any extra training supervision or modification to the optimization
 process and training images. We show that random transformation provides significant improvements of CNNs on many
 benchmark tasks, including small-scale image recognition, large-scale image
 recognition, and image retrieval. The code is available at
 \url{https://github.com/jasonustc/caffe-multigpu/tree/TICNN}.

\end{abstract}

%
%
%

%
%

%
%
\printccsdesc


\keywords{Convolutional Neural Networks; transform invariance.}

\section{Introduction}
In recent years, computer vision has been significantly advanced by the adoption of
convolutional neural networks (CNNs). We are currently witnessing many CNN-based models achieving
state-of-the-art results in many vision tasks, including image recognition \cite{GoogleNet, VGG, ResNet, gan2016learning, zhangyang-mm}, semantic segmentation \cite{FCNN}, image captioning \cite{Feifei-Image-Cap, Jeff-Image-Cap, MS-Image-Cap}, image retrieval \cite{zhengliang-cvpr, Zheng2016, zheng2014packing}, action recognition \cite{action-rcnn, action-two-stream}, video concept learning \cite{gan2015devnet, gan2016you} and video captioning \cite{yaoli-video-caption, msra-video-caption}.

\begin{figure*}[htb]
\begin{center}
\subfigure[CNN]{
\includegraphics[width=0.45\textwidth]{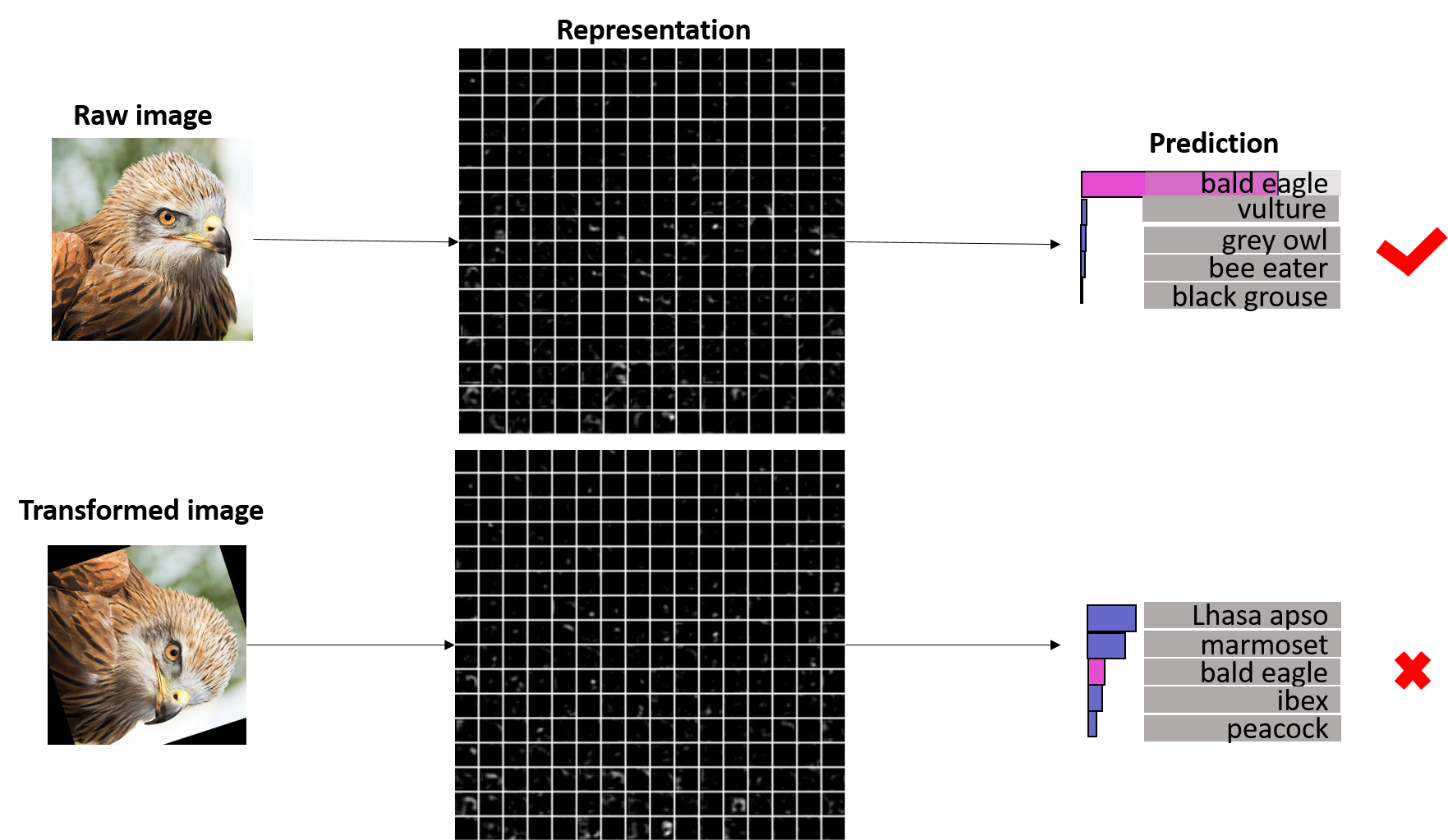}}
\hspace{5mm}
\subfigure[Transform Invariant CNN]{
\includegraphics[width=0.45\textwidth]{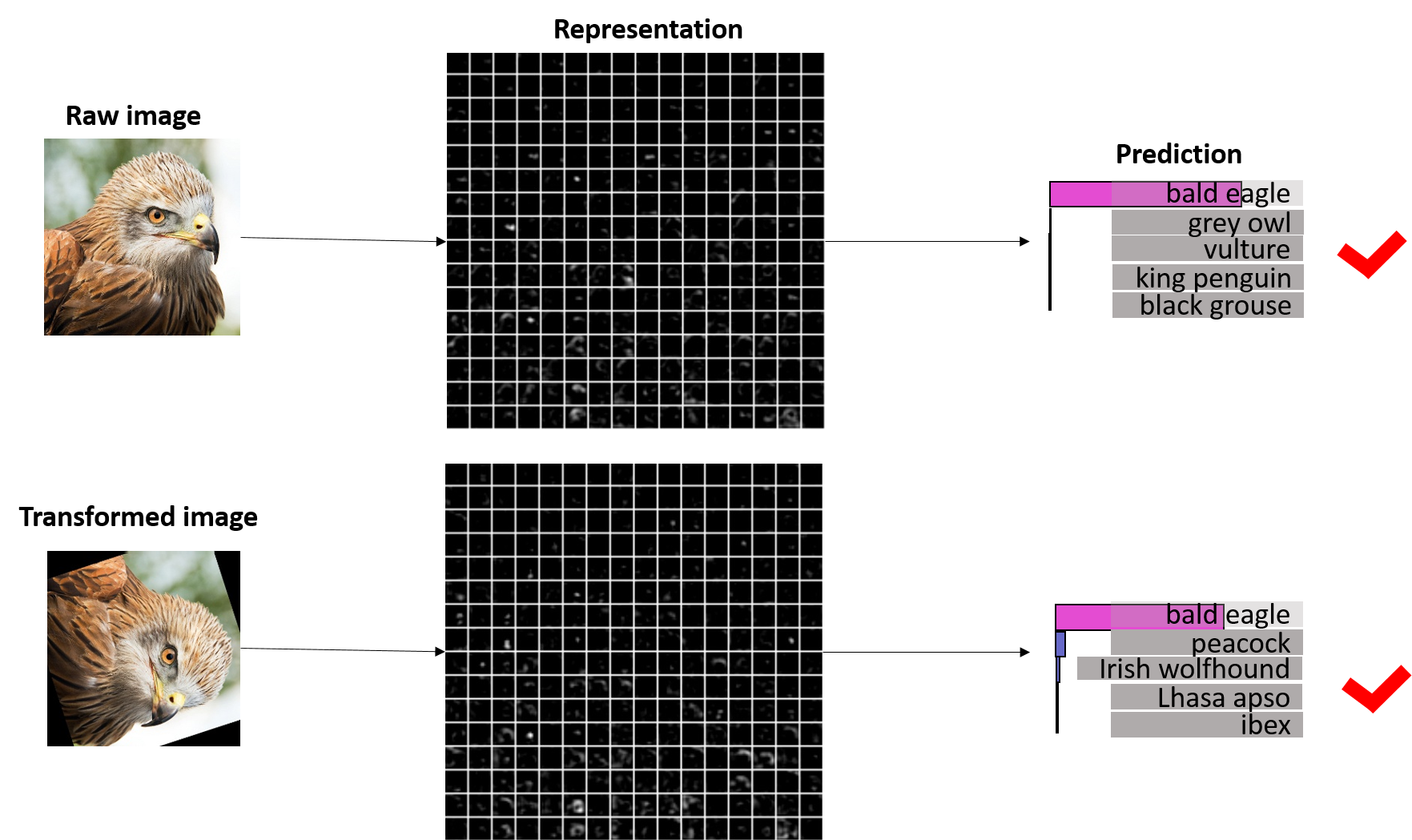}}
\caption{Limitation of current CNN models. The length of the
  horizontal bars is proportional to the probability assigned to the
  labels by the model, and pink indicates ground truth. Transform of
  input image causes the CNN to produce an incorrect prediction
  (a). Additionally, the representations ($256\times14\times14$ conv5 feature maps of AlexNet \cite{AlexNet}) are quite different,
while the representation and prediction of our transform-invariant CNN (same architecture as \cite{AlexNet}) is more
consistent (b).}
\label{fig:cnn-example}
\vspace{-6mm}
\end{center}
\end{figure*}

The local transform-invariant property of CNNs lies in the combination of local receptive fields with shared weights and
pooling. Because distortions or shifts of the input can cause the positions of salient features to vary, local receptive fields
with shared weights are able to detect invariant elementary features
despite changes in the positions of salient features
\cite{Lecun-cnn}. Moreover, average-pooling or max-pooling reduces the resolution of the feature maps in each layer,
which reduces the sensitivity of the output to small local shifts and distortions. However, due to the typically small
local spatial support for pooling (e.g., $2\times2$ pixels) and convolution (e.g., $9\times9$ kernel size), large global
invariance is only possible for a very deep hierarchy of pooling and convolutions, and the intermediate feature maps in
CNNs are not invariant to large transformations of the input data \cite{cnn-invariance}. This limitation of CNNs results from
the poor capacity of pooling and the convolution mechanism in learning global transform-invariant representations
(Fig. \ref{fig:cnn-example}).

In this work, we introduce a random transformation module of feature maps that can be included in any layer
in a CNN. Intuitively, with sufficient layers and parameters, hierarchical combinations of convolution
(matrix multiplication and non-linear activation) and pooling operations should have sufficient complexity to learn a
robust mapping from input images with any transform to transform-invariant representations. All we need is to introduce
a global way to push the CNN to learn with little dependency on the transform of input images. Our method is inspired by the success of
dropout \cite{dropout}. In dropout, each hidden unit is randomly omitted
from the network with a probability (e.g., $0.5$) on each presentation of each training case.
Therefore, a hidden unit cannot rely on other hidden units being present.
In this way, each neuron learns to detect a independent feature that is generally robust for producing the correct answer with
little dependency on the variety of internal contexts in the same layer. Similarly, randomly transforming
 (rotation, scale, and translation) feature maps of CNNs during the training stage prevents complex
 dependencies of specific rotation, scale, and translation levels of training images in CNN models. Rather,
 each convolutional kernel learns to detect a feature that is generally helpful for producing the
 transform-invariant answer given the combinatorially large variety of transform levels of its input
 feature maps.

In contrast to pooling layers, in which receptive fields are fixed and local, the random transformation is performed on the entire
feature map (non-locally) and can include any transformation, including
scaling, rotation, and translation. This guides the CNN to learn global transformation-invariant representations from raw input images. Notably, CNNs with random transformation layers
can be trained with standard back-propagation, allowing for end-to-end
training of the models in which they are injected. In addition,
we do not require any extra training supervision or modification of the optimization process or any transform of training images.

The main contributions of this work are as follows. (i) We propose a very simple approach to train robust transform-invariant convolutional neural networks. (ii) Because our model does not possess any extra parameters or extra feature extraction modules, it can be used to replace traditional CNNs in any
CNN-based model. Therefore, this is a general approach for improving the performance of CNN-based models in any vision task.
(iii) The validity of our method confirms that CNNs still have the potential to model a more robust mapping from transform-variant images to transform-invariant representations; we just need to push them to do so.

The remainder of this paper is structured as follows. Related works of transform-invariant CNNs are discussed in Section \ref{sec:related_work}.
The architecture and learning procedure of the proposed transform-invariant CNNs are described in Section \ref{sec:TICNN}. Extensive evaluations
compared with the state-of-the-art and comprehensive analysis of the proposed approach are reported in Section \ref{sec:exp}.
Conclusions are presented in Section \ref{sec:conclusion}.

\section{Related Work} \label{sec:related_work}
The equivalence and invariance of CNN representations to input image transformations were investigated in  \cite{cvpr-invest-inva, iclr-invest-inva, nips-invest-inva}. Specifically, Cohen and Welling \cite{iclr-invest-inva}
showed that the linear transform of a good visual representation was equivalent to a combination of the elementary
irreducible representations by using the theory of group representations. Lenc and and Vedaldi \cite{cnn-invariance}
estimated the linear relationships between representations of the original and transformed images. Gens and Domingos
\cite{nips-invest-inva} proposed a generalization of CNN that forms feature maps over arbitrary symmetry groups
based on the theory of symmetry groups in \cite{cvpr-invest-inva}, resulting in feature maps that were more invariant
to symmetry groups. Bruna and Mallat \cite{pami-invariant} proposed a wavelet scattering network to compute a translation-invariant image
representation. Local linear transformations were adopted in feature learning algorithms in \cite{icml-local-trans}
for the purpose of transformation-invariant feature learning.

Many recent works have focused on introducing transformation invariance in deep learning architectures explicitly.
For unsupervised feature learning, Sohn and Lee \cite{icml-local-trans} presented a transform-invariant restricted Boltzmann
machine that compactly represented data by its weights and their transformations, which achieved invariance of the feature
representation via probabilistic max pooling. Each hidden unit was augmented with a latent transformation assignment variable
that described the selection of the transformed view of the weights associated with the unit in \cite{inva-rbm}. In these
two works, the transformed filters were only applied at the center of the largest receptive field size. In tied convolutional
neural works \cite{tiled-cnn}, invariances were learned explicitly by square-root pooling hidden units that were computed by
partially un-tied weights. Here, additional learned parameters were needed when un-tying weights.
Alvarez \emph{et al.} \cite{fuse-cnn} learned each ConvNet over multiple scales independently without weight sharing and
fused the outputs of these ConvNets.  Sermanet and LeCun
\cite{multi-scale-cnn} utilized multi-scale CNN features by feeding
the outputs of all the convolutional layers to the classifier. In these two models, multiple-scale information was captured from different levels or
models of the hierarchy, but scale invariance was not encoded in the features learned at a layer because each layer is only
applied to the original scale. In \cite{lecun-hier-cnn}, the raw input image was transformed through a Laplacian pyramid. Each
scale was fed into a 3-stage convolutional network, which produces a set of feature maps with different scales disjointly.
Then, the outputs of all scales were aligned by up-sampling and
concatenated. However, concatenating multi-scale outputs to extract
scale-independent features involves extra convolution models and extra
parameters, thereby increasing the complexity of the models and the computational cost.

The latest two works on incorporating transform invariance in CNNs are described in \cite{sci-cnn, spatial-trans-cnn}.
In \cite{sci-cnn}, feature maps in CNNs were scaled to multiple levels, and the same kernel was convolved across the input in each
scale. Then, the responses of the convolution in each scale were normalized and pooled at each spatial location to obtain a locally
scale-invariant representation. In this model, only limited scales were considered, and extra modules were needed in the feature extraction process. To address different transformation types in input images, \cite{spatial-trans-cnn} proposed inserting
a spatial transformer module between CNN layers, which explicitly
transformed the input image into the proper appearance and fed that
transformed input into CNN model. In this work, invariance comes from
the extra transformation module, while invariance of the CNN itself is not improved.

In conclusion, all the aforementioned related works improve the transform invariance of deep learning models by adding
\textit{extra feature extraction modules}, \textit{more learnable parameters} or \textit{extra transformations on input images},
which makes trained CNN models problem dependent and not easy to be
generalized to other datasets. In contrast, in this paper, we propose
a very simple random transform operation on feature maps during the
training of CNN models. Intrinsic transform invariance of the current CNN model is obtained by pushing the model to learn more robust parameters from raw input images only.
No extra feature extraction modules or more learnable parameters are required.
Therefore, it is very easy to apply the trained transform-invariant CNN model to any vision task because we only need to replace current CNN parameters by this trained model.

\section{Transform-Invariant Convolutional Neural Networks} \label{sec:TICNN}
\subsection{Convolutional Neural Networks}
Convolutional neural networks (CNNs) are a supervised feed-forward multi-layer architecture in which each layer learns
a representation by applying feature detectors to previous feature maps. The outputs of the final layer are fed into
a classifier or regressor with a target cost function such that the network can be trained to solve supervised tasks.
The entire network is learned through stochastic gradient descent with gradients obtained by back-propagation \cite{Lecun-cnn}.
In general, a layer in a CNN consists of local patch-based linear feature extraction and non-linear activation, occasionally followed
by spatial pooling or normalization.

The core concept of CNNs is a combination of local receptive fields, shared weights, and spatial pooling. Each unit in a layer
receives inputs from a set of units located in a small neighborhood in the previous layer. This idea of connecting units
to local receptive fields is inspired by Wiesel's discovery of locally
sensitive and orientation-selective neurons in the
visual system of cats \cite{Lecun-cnn}. The output of convolving an input layer with one kernel is called a \textit{feature map}.
Because elementary feature detectors (convolutional weights) that are useful on one part of the
image should also be useful across the entire image, units in the same feature map share the same set of weights at multiple
locations. This is a good way to reduce the number of trainable parameters and ensure some local shift and distortion
invariance. Pooling is a simple way to reduce the spatial resolution of the feature map and also reduces the
sensitivity of convolution outputs to shifts and distortions.

In the forward pass of a CNN, each output feature map is computed by convolving input feature maps with kernel weights and then
activated by a non-linear activation function:
\begin{equation}
x^l_j = f((W_j^l*x^{l-1}) + b_j^l)
\end{equation}
where $*$ is the convolution operator, $x^{l-1}$ is the input feature maps from the previous layer, and $f$ is a non-linear function.
$W_j^l$ and $b_j^l$ are the trainable weights and bias of $j$th output feature map $x_j^l$.

\subsection{Transformation of an Image}
The basic $2$D transformation of an image consists of $3$ types: \textit{translation}, \textit{rotation}, and \textit{scale} \cite{computer-vision}.
The locations of pixels in an image before transformation and after transformation are
denoted as $(x,y)$ and $(x',y')$, respectively. Translation is calculated by:
\begin{equation}
\label{translation}
 (x',y',1) = (x, y, 1) \left[ \begin{array}{ccc}
1 & 0 & 0 \\
0 & 1 & 0 \\
d_x & d_y & 1 \\ \end{array} \right]
\end{equation}
where $d_x$ and $d_y$ are the numbers of translated pixels in the $x$
and $y$ axes, respectively.

Scale is computed by:
\begin{equation}
\label{scale}
 (x',y',1) = (x, y, 1) \left[ \begin{array}{ccc}
s_x & 0 & 0 \\
0 & s_y & 0 \\
0 & 0 & 1 \\ \end{array} \right]
\end{equation}
where $s_x$ and $s_y$ are the scale factors in the $x$ and $y$ axes, respectively.

Rotation is computed by:
\begin{equation}
\label{rotation}
 (x',y',1) = (x, y, 1) \left[ \begin{array}{ccc}
cos\theta & sin\theta & 0 \\
-sin\theta & cos\theta & 0 \\
0 & 0 & 1 \\ \end{array} \right]
\end{equation}
where $\theta$ is the rotation angle ranging from $-180^\circ$ to $180^\circ$.

When translation, scale, and rotation are applied to an image simultaneously, the output can be computed as:
\begin{equation}
\label{all}
\begin{aligned}
 (x',y',1) = (x, y, 1) \left[ \begin{array}{ccc}
cos\theta & sin\theta & 0 \\
-sin\theta & cos\theta & 0 \\
0 & 0 & 1 \\ \end{array} \right] \\
\left[ \begin{array}{ccc}
s_x & 0 & 0 \\
0 & s_y & 0 \\
0 & 0 & 1 \\ \end{array} \right] \\
\left[ \begin{array}{ccc}
1 & 0 & 0 \\
0 & 1 & 0 \\
d_x & d_y & 1 \\ \end{array} \right]
\end{aligned}
\end{equation}

\subsection{Transform-Invariant Convolutional Neural Networks}

\begin{figure*}[t!]
\centering
\includegraphics[width=0.80\textwidth]{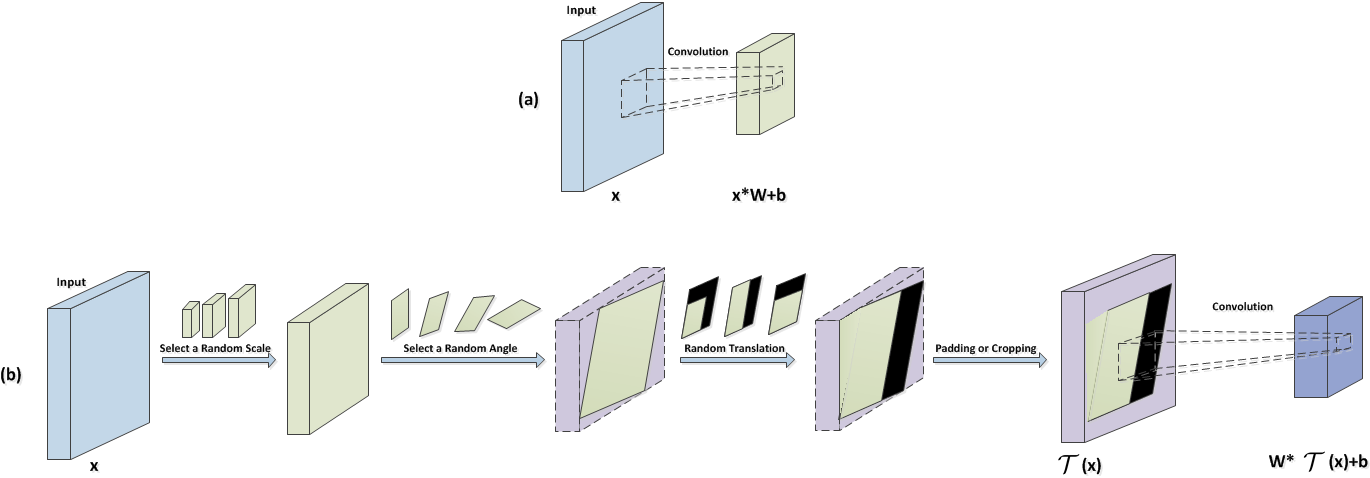}
\caption{Detailed comparison of the structure of (a) convolution layer and the proposed (b) \textit{transform-invariant} convolution layer. In (b), after convolving the inputs with the kernel, the output feature maps
are transformed by a random rotation angle, scale factor, and translation proportion. Then, the randomly transformed
feature maps are fed into the next layer.}
\label{fig:compare}
\end{figure*}

Weight sharing makes CNN capable of detecting features regardless of their spatial locations in the input feature maps.
Pooling reduces the sensitivity of CNN to local shifts and distortions in input images. However, when faced with global
transformations of input images, the performance of CNNs is significantly decreased. In this section, we describe our
transform-invariant CNN (TICNN). By introducing random transformations
(scale, rotation, and translation) of feature maps in the CNN model, convolutional
kernels are trained to be invariant to these
transformations. Fig. \ref{fig:compare} presents a side-by-side comparison of the overall structure of these two CNN layers.

One perspective to interpret TICNN is to compare it with dropout \cite{dropout}.
Dropout attempts to introduce a neuron-level randomness to deep
models; in this way, the dependency of the feature detector on the existence of neurons in the same layer is reduced.
TICNN attempts to incorporate a feature map (concept)-level randomness in CNN models, which can reduce the dependence of feature detectors
in the current layer on the transform level of previous feature maps (appearance).

From the perspective of transformation, Jaderberg \emph{et al}. \cite{spatial-trans-cnn} proposed transforming the input into a proper and easy-to-recognize
appearance. TICNN forces the architecture to learn a more robust model for all
possible transformations, and the intrinsic transform invariance of the architecture is improved.

From the perspective of data augmentation, training data augmentation and multi-scale model augmentation \cite{sci-cnn, lecun-hier-cnn} only
provide the model with limited transform levels of input images or
feature maps. While in TICNN, as training proceeds, the model will
learn almost all transformation levels of inputs randomly. Therefore, TICNN should be more robust when input images are transformed.

Considering the spatial range of transformation invariance, pooling
can only address a local invariance for shifts and distortions
(e.g., $3\times3$ pixels). Our TICNN paves the way for global invariance of transformations in CNN models.

\subsubsection{Forward Propagation}
Our goal is to push each feature detector to learn a feature that is generally helpful for producing the transform-invariant
answer given a large variety of transform levels of its input feature maps.
To accomplish this goal, before being convolved by the convolutional
kernels/filters, the input feature maps are first scaled, rotated and
translated via a random sampled scale factor, rotation angle and translate proportion.
To keep the size of feature maps unchanged, the transformed feature maps are either cropped or padded with $0$s to be
properly convolved.

Specifically, let $\mathcal{T}$ be a linear image transform operator that applies random spatial transformations to an input $x$. The output feature map $y$ is computed as:
\begin{equation}
\hat{x} = \mathcal{T}(x)
\end{equation}
\begin{equation}
y = f((W*\hat{x}) + b)
\end{equation}

Here, the $3\times3$ transformation matrix $\mathcal{T}$ is randomly sampled as follows:
\begin{equation}
\label{all1}
\begin{aligned}
 \mathcal{T} = \left[ \begin{array}{ccc}
cos\theta & sin\theta & 0 \\
-sin\theta & cos\theta & 0 \\
0 & 0 & 1 \\ \end{array} \right] 
\left[ \begin{array}{ccc}
s_x & 0 & 0 \\
0 & s_y & 0 \\
0 & 0 & 1 \\ \end{array} \right] 
\left[ \begin{array}{ccc}
1 & 0 & 0 \\
0 & 1 & 0 \\
d_x & d_y & 1 \\ \end{array} \right]
\end{aligned}
\end{equation}
where $\theta\sim\mathcal{N}(\mu_\theta, \sigma_\theta^2)$, $s_x, s_y \sim\mathcal{N}(\mu_s, \sigma_s^2)$, and
$d_x, d_y \sim\mathcal{N}(\mu_d, \sigma_d^2)$. In the presence of each training image, \{$\theta$, $s_x$, $s_y$,
$d_x$, and $d_y$\} are randomly sampled.

\subsubsection{Back-Propagation}

Because the transform-invariant convolution layer consists of only linear operations, its gradients can be computed
by a simple modification of the back-propagation algorithm. For each transformation, the error signal is propagated
through the bilinear coefficients used to compute the transformation.

Let $x^{l-1}$ denote the vectorized input of layer $l$ of length $m$ ($m=hwc$ for a $h$ by $w$ by $c$ image). If the
dimensionality of the transformed output is $n$, the transformation $\mathcal{T}$ of input $x^{l-1}$ can be computed via $Tx^{l-1}$, where $T$ is an $n\times m$ interpolation coefficient matrix.
Specifically, each row of $T$ has $4$ non-zero coefficients with bilinear interpolation.
By encoding kernel weights $W$ as a Toeplitz matrix, the convolution operation can be expressed as:
\begin{equation}
z^l = toep(W)(Tx^{l-1}) + b^l
\end{equation}
\begin{equation}
x^l = f(z^l)
\end{equation}
where $f$ is a non-linear function and $b^l$ is the bias.

Then, the error propagated from the previous layer $\delta^{l+1}$ to
the current layer $\delta^l$ can be computed by:
\begin{equation}
\delta^l = T^{\top}toep(W)^{\top}f'(z^l)\odot \delta^{l+1}
\end{equation}

\subsubsection{Testing}
Because our model does not introduce any extra parameters or feature extraction modules, we only need to replace
parameters in the CNN model with parameters trained by our model. The test architectures and all other settings are
identical.

\section{Experiments} \label{sec:exp}
In this section, we evaluate our proposed transform-invariant CNN on
several supervised learning tasks. First, to compare our model with
state-of-the-art methods, we conduct experiments on the distorted
versions of the MNIST
handwriting dataset as in \cite{sci-cnn, spatial-trans-cnn}. The results in Section \ref{sec:mnist} show that TICNN is capable
 of achieving comparable or better classification performance with only raw input images and without adding any extra
feature extraction modules or learnable parameters. Second, to test the improvement of TICNN on CNN for large-scale
real-world image recognition tasks, we compare our model with AlexNet
\cite{AlexNet} on the ImageNet-2012 dataset in Section
\ref{sec:imagenet}. The results demonstrate that, with little
degradation on the performance of classifying original test images,
random transform can be applied to any layer in the CNN model and is able to consistently improve the performance on
transformed images significantly. Finally, to evaluate the
generalization ability of the TICNN model on other vision tasks
with real-world transformations of images, we apply our model to solve
the image retrieval task on the UK-Bench \cite{uk-bench} dataset in
Section \ref{sec:uk-bench}. The improvement of retrieval performance
reveals that the TICNN model has a good
generalization ability and is better at solving real-world transformation varieties. We implement our method using the
open source Caffe framework \cite{caffe}. Our code and model will be available online.

\begin{figure}
\centering
\includegraphics[width=0.35\textwidth]{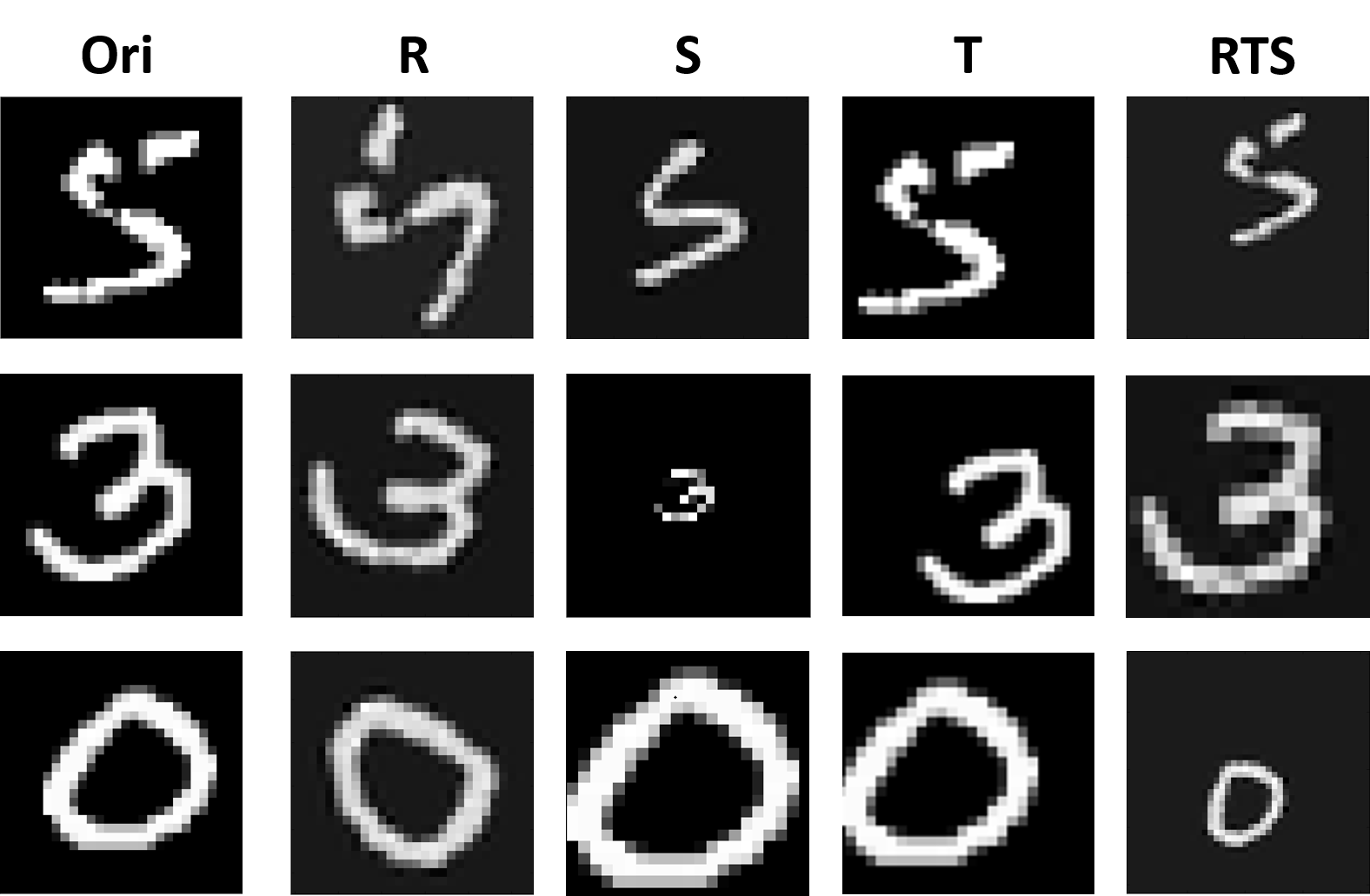}
\caption{Examples of distorted MNIST dataset. Ori: raw images. R: rotated images. S: scaled images. T: translated images.
 RTS: images with scale, rotation and translation. }
 \label{fig:sample-mnist}
\end{figure}

\subsection{MNIST}\label{sec:mnist}
In this section, we use the MNIST handwriting dataset to evaluate all deep models. In particular, different neural networks
are trained to classify MNIST data that have been transformed in various ways, including rotation (R), scale (S), translation (T), and rotation-scale-translation (RTS). The rotated dataset was generated from rotating MNIST digits with a random angle sampled from a uniform distribution
$U[-90^{\circ}, 90^{\circ}]$. The scaled dataset was generated by
scaling the digits randomly with a scale factor sampled from $U[0.7,
  1.2]$. The translated dataset was generated from randomly locating
the $28\times28$ digit in a $42\times42$ canvas. The
rotated-translated-scaled dataset (RTS) was generated by randomly
rotating the digit by $-45^{\circ}$ to $-45^{\circ}$, randomly scaling
the digit by a factor ranging from [$0.7$, $1.2$] and placing the
digit at a location in a $42\times42$ image randomly, all with a uniform distribution. Some example images are presented in Fig.
\ref{fig:sample-mnist}.

\begin{table}[t]
\begin{center}
\begin{tabular}{|l|c|c|c|c|}
\hline
Method & R & S & T & RST \\
\hline
FCN & 2.1 & 2.8 & 2.9 & 5.2 \\
\hline
CNN & 1.2 & 1.4 & 1.3 & 0.8 \\
\hline
ST-CNN\cite{spatial-trans-cnn} & \textbf{0.8} & 0.9 & 0.8 & \textbf{0.6} \\
\hline
SI-CNN\cite{sci-cnn} & - & 1.0 & - & - \\
\hline
TI-CNN(ours) & \textbf{0.8} & \textbf{0.8} & \textbf{0.7} & \textbf{0.6} \\
\hline
\end{tabular}
\end{center}
\caption{Classification error on the transformed MNIST dataset. The different distorted MNIST datasets
are R: rotation, S: scale, T: translation, and RST: rotation-scale-translation. All models
have the same number of parameters and use the same basic architectures.}
\label{perf:mnist}
\end{table}

\begin{figure}
\centering
\includegraphics[width=0.45\textwidth]{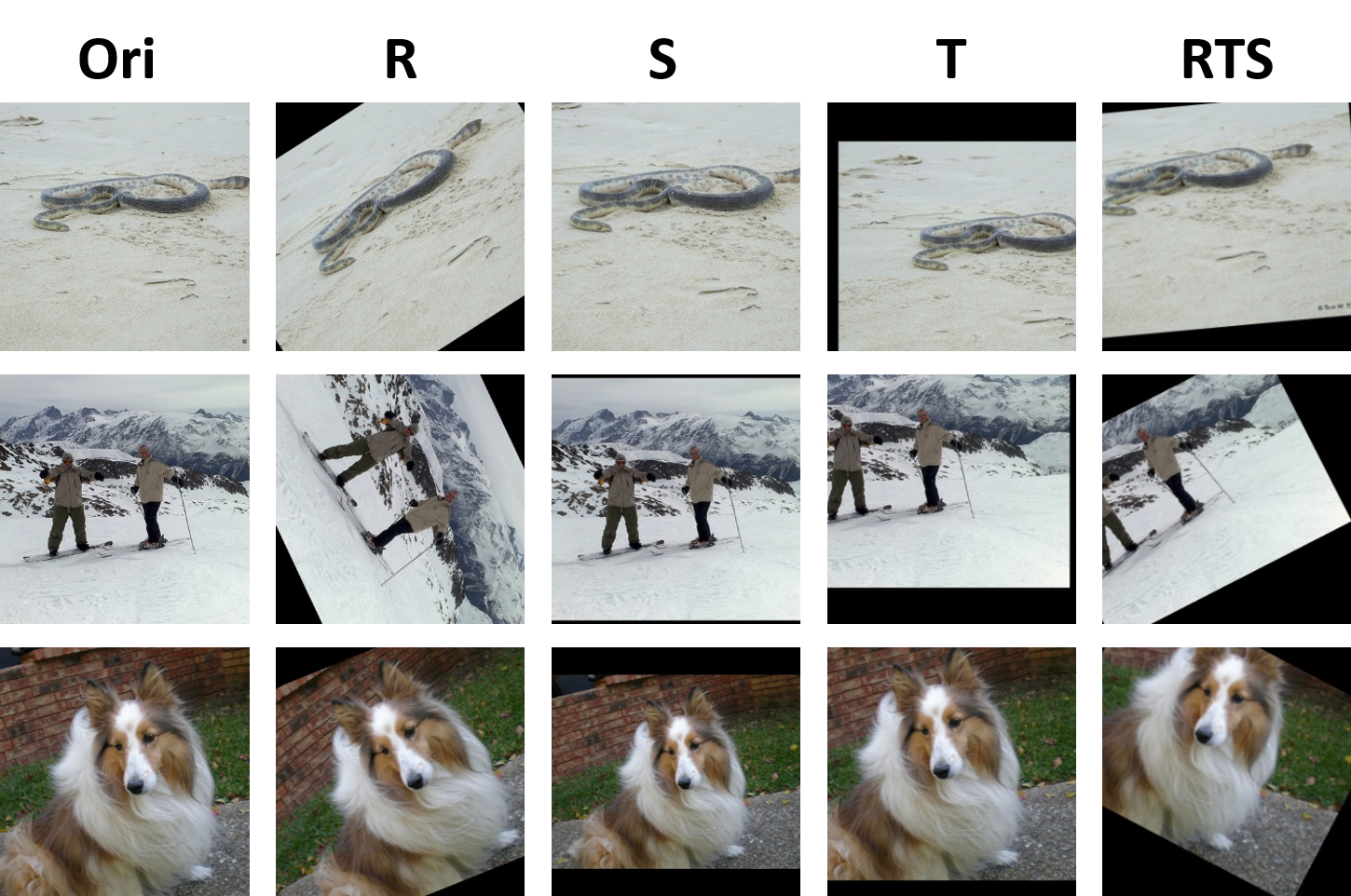}
\caption{Distorted images in ImageNet. Ori: raw images. R: rotated images. S: scaled images. T: translated images.
 RTS: images with scale, rotation and translation.}
 \label{fig:sample-imagenet}
\end{figure}

Following \cite{spatial-trans-cnn}, all networks use the ReLU activation function and softmax classifiers. All CNN networks have a $9\times9$ convolutional layer (stride $1$, no padding), a $2\times2$ max-pooling layer with stride $2$, a subsequent $7\times7$ convolutional layer (stride $1$, no padding), and another $2\times2$ max-pooling layer with
stride $2$ before the final classification layer. All CNN networks
have $64$ filters per layer. For the method in \cite{sci-cnn}, convolution layers are replaced by scale-invariant layers using six scales from $0.6$ to $2$ at a scale
step of $2^{1/3}=1.26$. The spatial transformer module is placed
at the beginning of the network \cite{spatial-trans-cnn}. In our random transform-invariant CNN, the random transformation
of the feature map module is applied to the second convolution layer,
and \textit{our model is only trained on the raw input images}. The
rotation angle is sampled from $\mathcal{N}(0, 30^2)$, the scale
factor is sampled from $\mathcal{N}(1, 0.15^2)$, and the translation proportion is sampled from $\mathcal{N}(0, 0.2^2)$.

All networks were trained with SGD for 150000 iterations, $256$ batch size, $0.01$ base learning rate, no weight decay, and no dropout.
The learning rate was reduced by a factor of $0.1$ every $50000$
iterations. The weights were initialized randomly,
and all networks shared the same random seed.

\begin{table*}[t]
\begin{center}
\begin{tabular}{|l|c|c|c|c|}
\hline
 Model & R  & T & S & RTS \\
 \hline
 CNN+Data Augmentation & 56.3/79.7 & 56.3/79.4 & 55.6/78.9 & 56.6/79.8 \\
\hline
TI-Conv1 & 55.0/78.4 & 56.6/79.7 & 56.2/79.1 & 56.5/79.4  \\
\hline
TI-Conv2 & 56.1/79.3 & 56.5/79.7 & 56.2/79.3 & 56.6/79.7  \\
\hline
TI-Conv3 & 56.7/79.6 & 56.9/79.9 & 56.4/79.4 & 57.0/80.0 \\
\hline
TI-Conv4 & 57.1/80.1 & 57.2/80.1 & 56.7/79.5 & 56.9/79.9 \\
\hline
TI-Conv5 & 57.1/80.2 & \textbf{57.6}/\textbf{80.5} & 57.0/80.0 & 57.1/80.0 \\
\hline
TI-Conv1,2,5 & 57.2/80.2 & 57.5/80.4 & 57.1/80.2 & 57.2/80.3 \\
\hline
SI-CNN & \multicolumn{4}{|c|}{57.3/80.3} \\
\hline
CNN & \multicolumn{4}{|c|}{57.1/80.2} \\
\hline
\end{tabular}
\end{center}
\vspace{-2mm}
\caption{Classification accuracy on the ILSVRC-2012 validation set (raw). TI-Conv$i$ indicates that we
apply random transformations on the feature maps of layer $i$. R, T, S, and RTS represent rotation, translation,
scale, and rotation-scale-translation, respectively.  }
\label{perf:imagenet-ori}
\end{table*}

\begin{table*}[t]
\begin{center}
\begin{tabular}{|l|c|c|c|c|}
\hline
 Model & R  & T & S & RTS \\
 \hline
CNN+Data Augmentation & 36.5/58.3 & 50.0/73.9 & 54.0/77.6 & 36.0/58.2 \\
\hline
TI-Conv1 & 35.7/57.9 & 50.5/74.4 & 54.0/77.5 & 35.6/58.0  \\
\hline
TI-Conv2 & 36.1/58.2 & 49.9/74.0 & 53.9/77.5 & 35.5/57.8  \\
\hline
TI-Conv3 & 36.8/58.9 & 50.0/74.2 & 54.6/77.9 & 36.3/58.9 \\
\hline
TI-Conv4 & 37.2/59.5 & 50.3/74.3 & 54.5/77.7 & 36.2/58.9 \\
\hline
TI-Conv5 & 37.1/59.2 & 50.1/74.6 & 54.7/78.0 & 36.0/58.7 \\
\hline
TI-Conv1,2,5 & \textbf{37.3/60.2} & \textbf{51.3/75.4} & \textbf{55.0/78.5} & \textbf{37.5/60.5} \\
\hline
SI-CNN & - & - & 53.8/77.3 & - \\
\hline
CNN & 36.6/57.7 & 46.5/70.8 & 53.3/76.9 & 33.3/55.2 \\
\hline
\end{tabular}
\end{center}
\vspace{-2mm}
\caption{Classification accuracy on the transformed ILSVRC-2012 validation set. TI-Conv$i$ indicates that we
apply random transformations on the feature maps of layer $i$. R, T, S and RTS represent rotation, translation, scale, and rotation-translation-scale, respectively.}
\label{perf:imagenet-trans}
\end{table*}

The experimental results are summarized in Table
\ref{perf:mnist}. Please notice that we do not report the best result of ST-CNN \cite{spatial-trans-cnn}, which was trained with a more narrow class of transformations selected manually (affine transformations).In our method, we did not optimize with respected to transformation classes. Therefore we compare with the most general ST-CNN defined for a class of projection transformations as in \cite{ti-pooling}. This table shows that our model achieves better performance on scale and
translation and comparable performance on rotation and rotation-scale-translation. Because our model does not require any extra learnable parameters, feature extraction modules, or transformations on training images,
the comparable performance still reflects the superiority of TICNN.
In addition, the experimental results also indicate that the
combination of convolution and pooling modules have some limitations
on learning the invariance of rotation. Placing extra transformation
modules at the beginning of the CNN is a more powerful way to compensate for this limitation.

\subsection{ILSVRC-2012}\label{sec:imagenet}
The ILSVRC-2012 dataset was used for ILSVRC $2012-2014$ challenges. This dataset
includes images of $1000$ classes, and it is split into three subsets: training ($1.3$M images),
validation ($50$K images), and testing ($100$K images with held-out class labels). The classification
performance is evaluated using two measures: the top-1 and top-5 errors. The former is a multi-class
classification error. The latter is the main evaluation criterion used in ILSVRC and is defined
as the proportion of images whose ground-truth category is not in the top-5 predicted categories.
We use this dataset to test the performance of our model on large-scale image recognition tasks.

CNN models are trained on raw images and tested on both raw and transformed images. For all transform types, specific
transformations are applied to the original images. Then, the
transformed images are rescaled to have a $256$ smallest image
side. Finally, the center $224\times224$ crop is used for the test. The rotated (R) dataset
was generated by randomly rotating the original images from
$-45^{\circ}$ to $45^{\circ}$ with a uniform distribution. The scaled
dataset (S) was generated by scaling the original images randomly in $U[0.7, 1.5]$. The translated dataset (T) was
generated by randomly shifting the image by a proportion of $U[-0.2,
  0.2]$. For the rotation-scale-translation (RTS)
dataset, all the aforementioned transformations were applied. Some examples of transformed images are presented in Fig.
\ref{fig:sample-imagenet}.

The architecture of our CNN model is the same as that of AlexNet \cite{AlexNet}. This model consists of $5$ convolution layers and
$2$ fully connected layers. The first convolution layer contains $96$ kernels of size $11\times11\times3$. The second
convolution layer contains $256$ kernels of size $5\times5\times48$. The third convolutional layer contains $384$ kernels of size
$3\times3\times256$. The fourth convolutional layer contains $384$
kernels of size $3\times3\times192$, and the fifth convolutional
layer contains $256$ kernels of size $3\times3\times192$. The fully connected layers have $4096$ neurons each. Please refer to
\cite{AlexNet} for more details.

The model is trained for $450,000$ iterations. We use a base learning rate of $0.01$ and decay it by $0.1$ every
$100,000$ iterations. We use a momentum of $0.9$, weight decay of $0.0005$, and a weight clip of $35$. The convolutional
kernel weights and bias are initialized by $\mathcal{N}(0, 0.01^2)$ and $0.1$, respectively. The weights and bias of
fully connected layers are initialized by $\mathcal{N}(0, 0.005^2)$ and $0.1$. The bias learning rate is set to be \
$2\times$ the learning rate for the weights.

To investigate the influence of a random transformation module on the CNN model, we apply it to different convolution layers.
The experimental results are summarized in Table \ref{perf:imagenet-ori} (test on raw validation set) and Table \ref{perf:imagenet-trans}
 (test on transformed validation set). Some key insights obtained from the results are listed as follows:
\begin{itemize}
    \item In general, when random transforms are applied to feature
      maps of higher layers in the CNN model, the recognition
    performance on both the original dataset and transformed dataset
    tends to be better. A possible explanation for this result is that
    representations extracted by lower layers are elementary features;
    they are related to a very precise location and
    have a very high resolution. Feature maps with higher resolution will be considerably more sensitive to transformations
    and noise, and location changes in lower-level feature maps could lead to a significant change in their visual
    pattern. Therefore, when a random transform is applied to feature maps of lower layers, it could be very difficult for the model
    to infer the original image content. Consequently, the useful information that the model can efficiently learn from each
    sample is reduced.

    \item By incorporating a random transformation, the model can still obtain comparable or even better
    performance than the CNN model on raw test images; this phenomenon shows that this randomness can prevent overfitting.
    Notably, the performance on transformed images is significantly improved
    on all transformation types (rotation, translation, scale, and rotation-translation-scale ), indicating that TICNN
    improves the generalization ability of the CNN model on transformation variances.

    \item Applying our random transform module on the image data layer is identical to unlimited data augmentation.
    Models with random transformations on the image data layer are worse than models with random transformations
    on the convolutional layer, revealing that TICNN model is better than simple image data augmentation.

    \item Comparing Table \ref{perf:imagenet-ori} with Table
      \ref{perf:imagenet-trans}, we can observe that the performance
    of CNN models significantly decreases when faced with
    transformations. The combination of convolution operations
    and pooling operations shows limited capacity for transform invariance. We need to add more powerful
    modules into current CNNs in the future.
\end{itemize}

\textbf{Effect on Feature Detectors.}
We evaluate the transform invariance achieved by our model using the invariance measure proposed in \cite{invariance-measure}.
In this approach, a neuron is considered to be firing when its response is above a certain threshold $t_i$. Each $t_i$ is chosen
to satisfy the condition that $G(i) = \sum |h_i(x) > t_i| / N$ is greater than $0.01$, where $N$ is the number of inputs.
Then \textit{local firing rate} $L(i)$ is computed as the proportion
of transformed inputs that a neuron fires to. To ensure that
a neuron is selective and with a high local firing rate (invariance to
the set of the transformed inputs), the invariance score of a
neuron is computed by the ratio of its invariance to selectivity, i.e., $L(i) / G(i)$. We report the average of the top $20\%$
highest scoring neurons ($p=0.2$) as in \cite{sci-cnn}. Please refer to \cite{invariance-measure} for more details.

\begin{figure}
\centering
\includegraphics[width=0.35\textwidth]{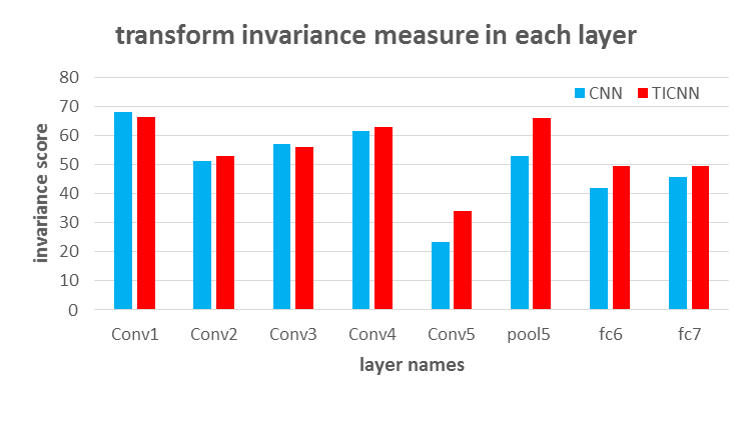}
\caption{Transform invariance measure (the larger the better).}
\label{fig:invariance-measure}
\end{figure}

Here, we construct the transformed dataset by applying rotation ($[-45^{\circ}, 45^{\circ}]$ with step size $9^{\circ}$), scale
($[0.7, 1.5]$ with step size $0.08$) and translation ($[-0.2, 0.2]$ with step size $0.04$) on the $50, 000$ validation images
of ImageNet-2012. Fig. \ref{fig:invariance-measure} shows the invariance scores of CNN (AlexNet) and TICNN (TI-Conv1,2,5) measured at the end of each layer. We can observe that by applying random transformation on feature maps during training, TICNN produces features that are more
transform invariant than those from CNN models.

\begin{figure*}
\centering
\includegraphics[width=0.65\textwidth]{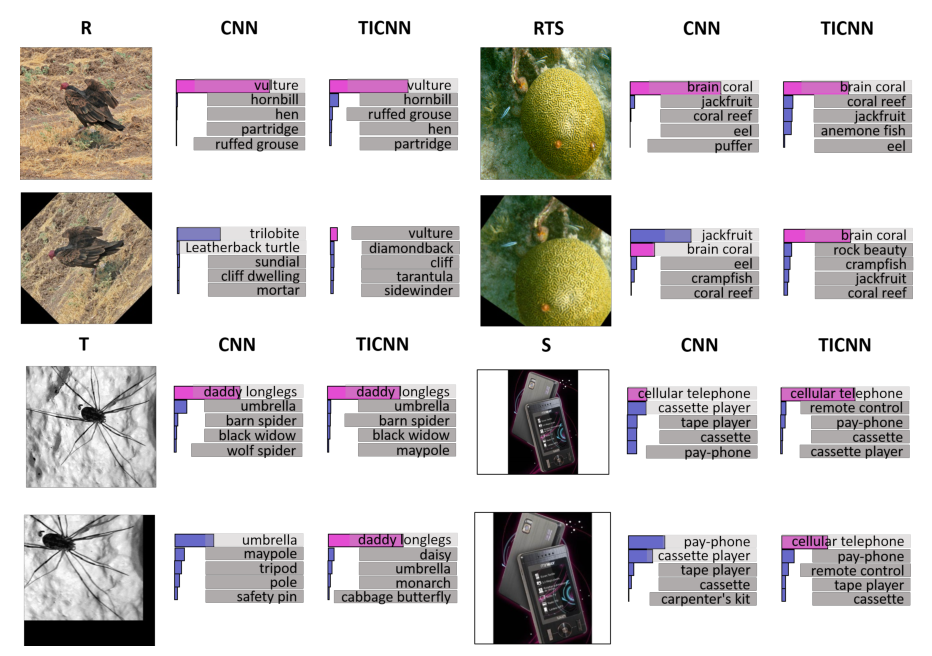}
\caption{Some ImageNet test cases with the $5$ most probable labels as predicted by AlexNet and our model.
The length of the horizontal bars is proportional to the probability assigned to the labels by the model.
Pink indicates ground-truth. The transformations applied to input images are rotation (R), scale (S),
translation (T), and rotation-scale-translation (RTS).}
\label{fig:imagenet-result}
\end{figure*}

\textbf{Some Examples.}
Figure \ref{fig:imagenet-result} shows some test cases in which CNN is
unable to predict correct labels for the input images while TICNN
generates true predictions. They demonstrate that TICNN is more robust
than the CNN model.

\subsection{UK-Bench}\label{sec:uk-bench}

\begin{figure*}[!t]
	\centering
	\subfigure[\label{fig:ill:1}]{\includegraphics[width=0.40\textwidth]{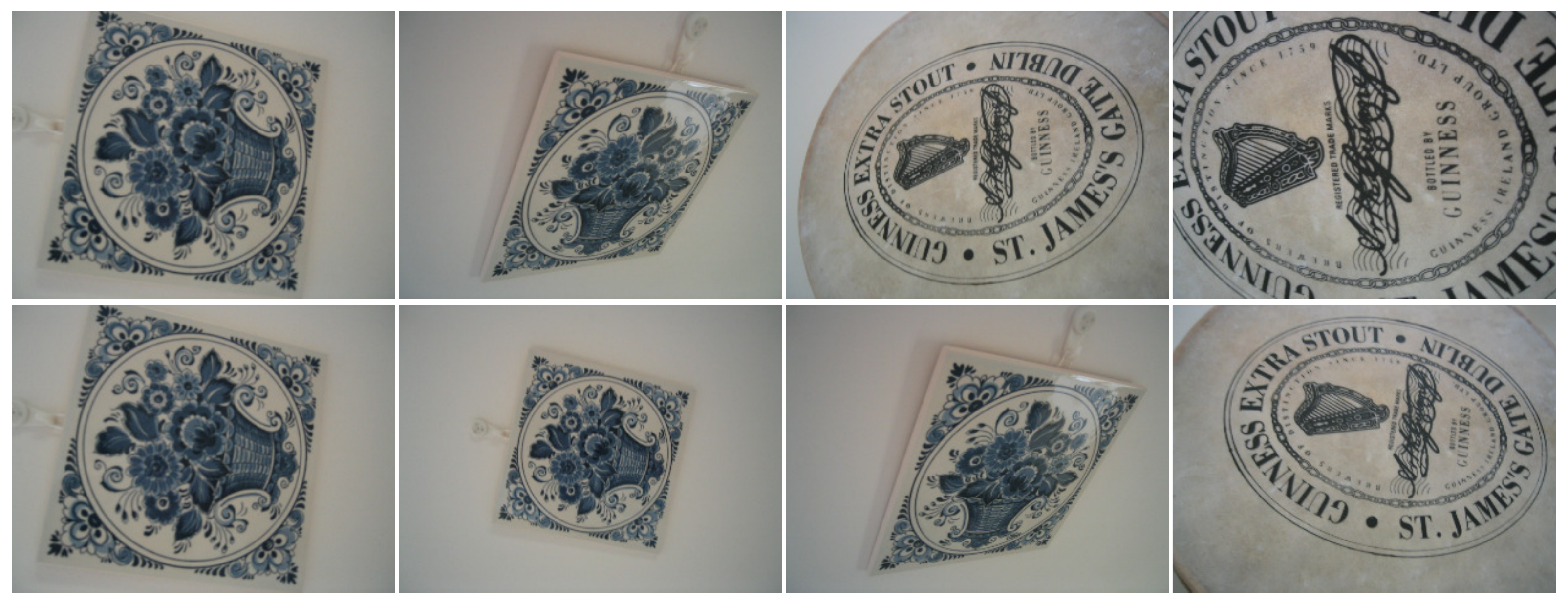}}
	\subfigure[\label{fig:ill:2}]{\includegraphics[width=0.40\textwidth]{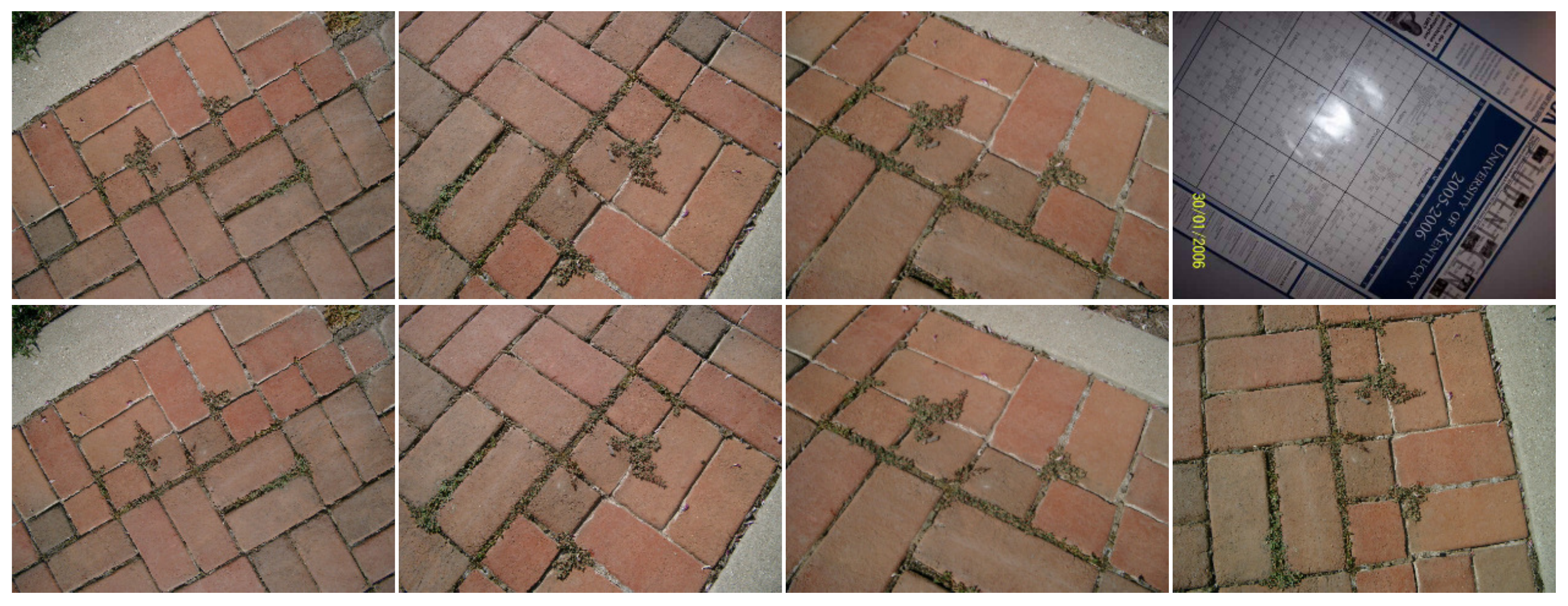}}\\
	\subfigure[\label{fig:ill:3}]{\includegraphics[width=0.40\textwidth]{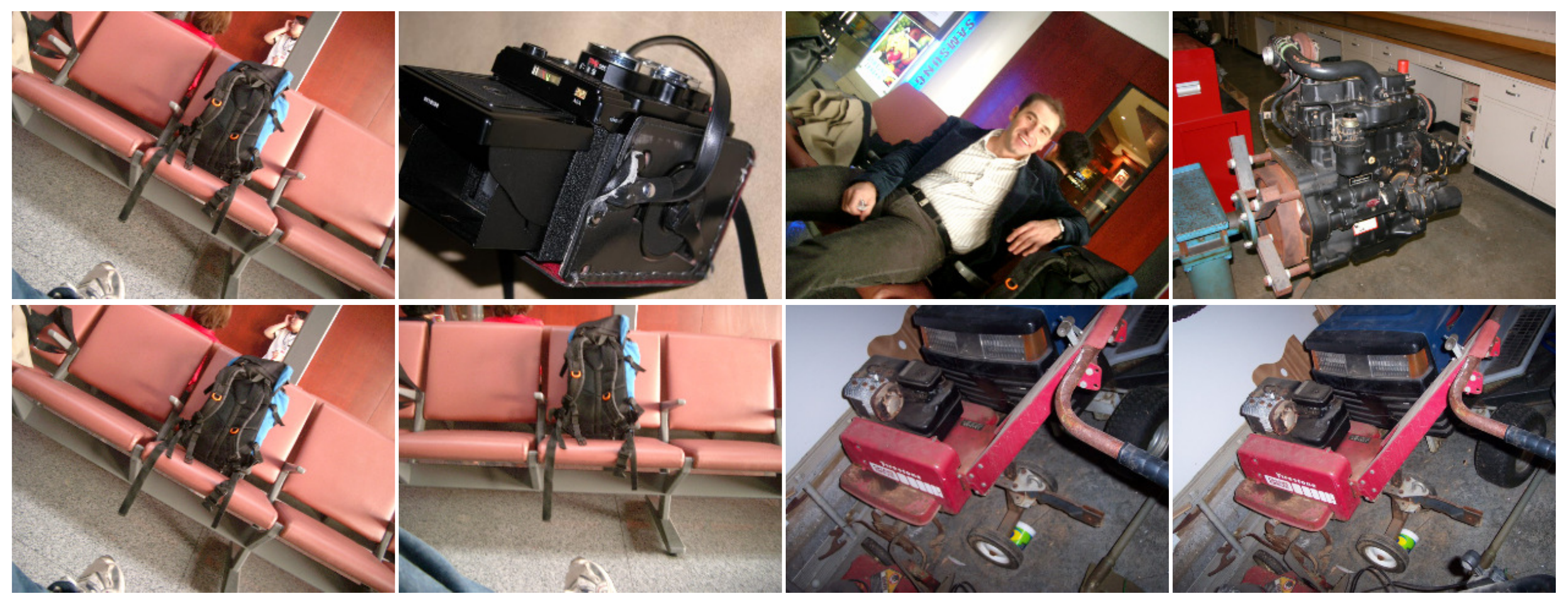}}
	\subfigure[\label{fig:ill:4}]{\includegraphics[width=0.40\textwidth]{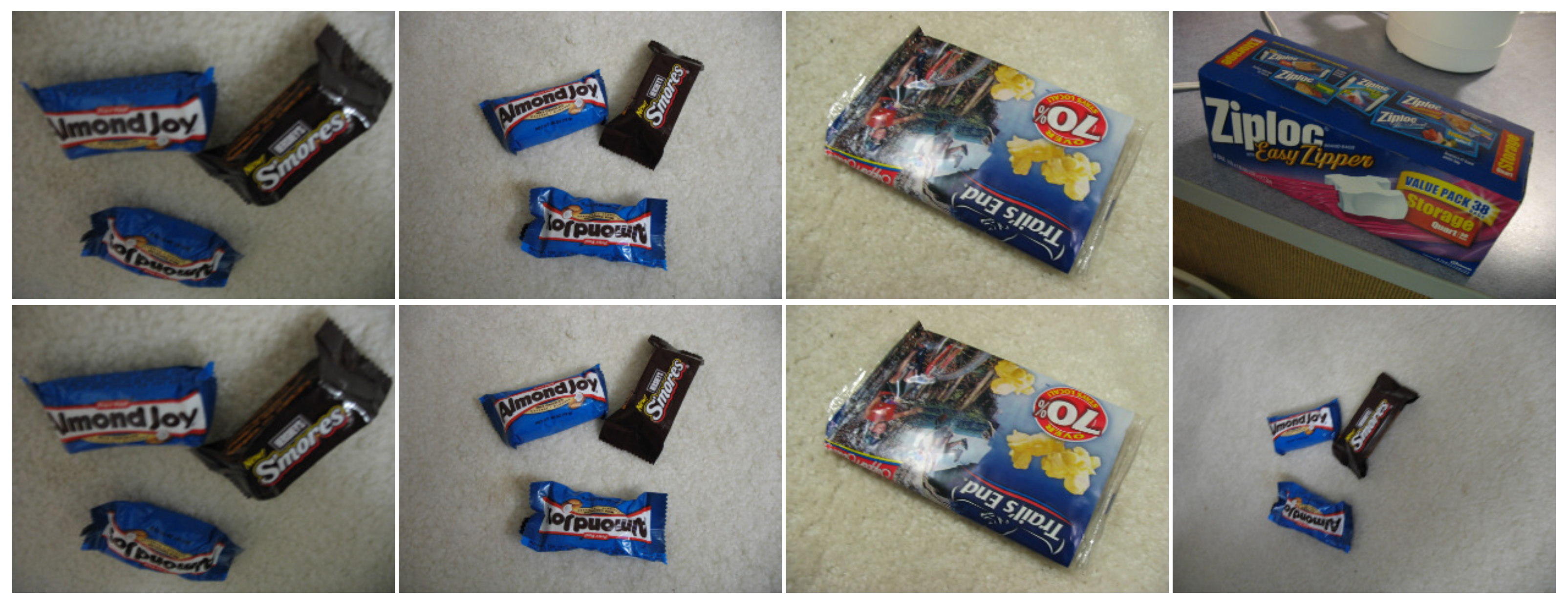}}\\
	\subfigure[\label{fig:ill:5}]{\includegraphics[width=0.40\textwidth]{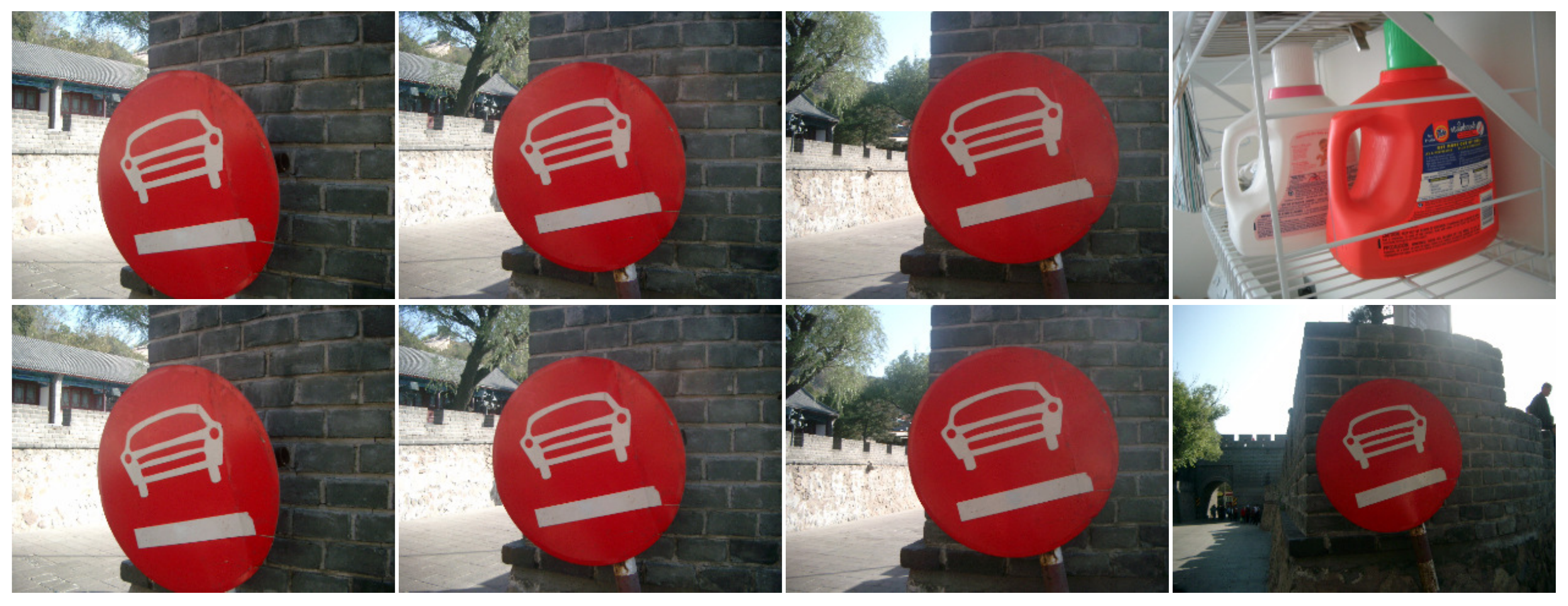}}
	\subfigure[\label{fig:ill:6}]{\includegraphics[width=0.40\textwidth]{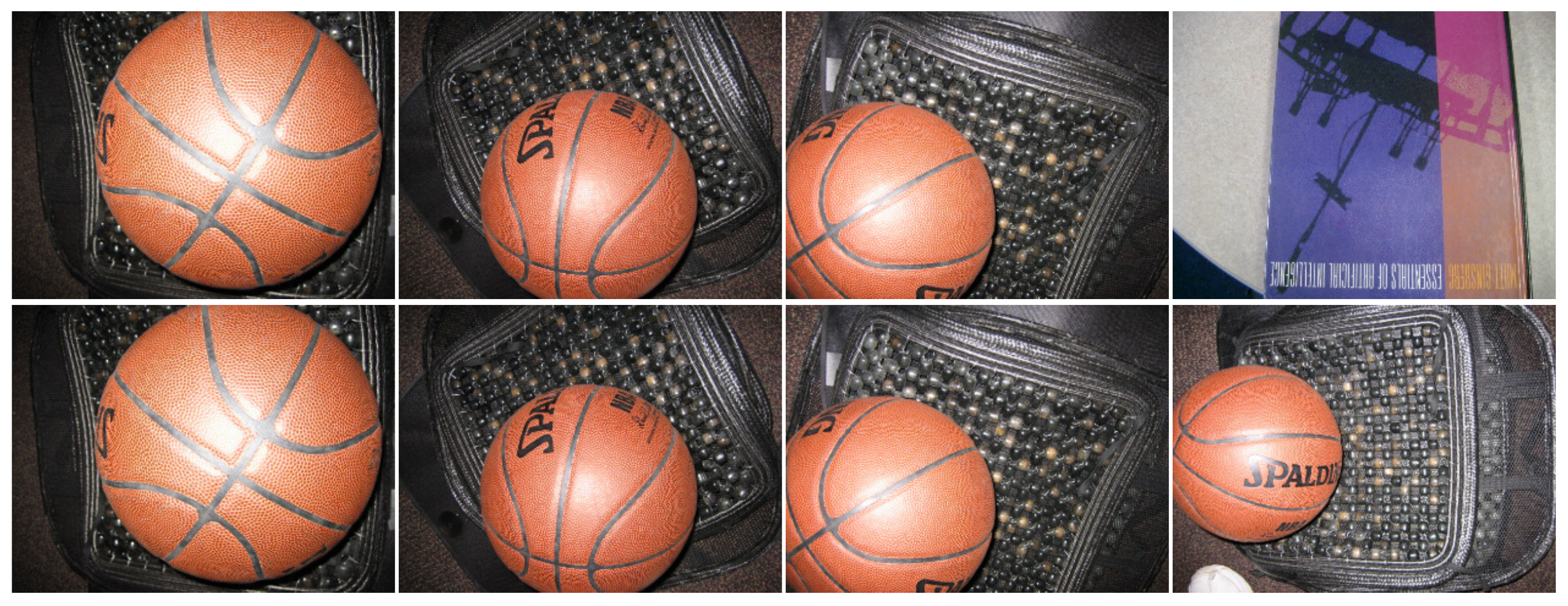}}\\
	\subfigure[\label{fig:ill:7}]{\includegraphics[width=0.40\textwidth]{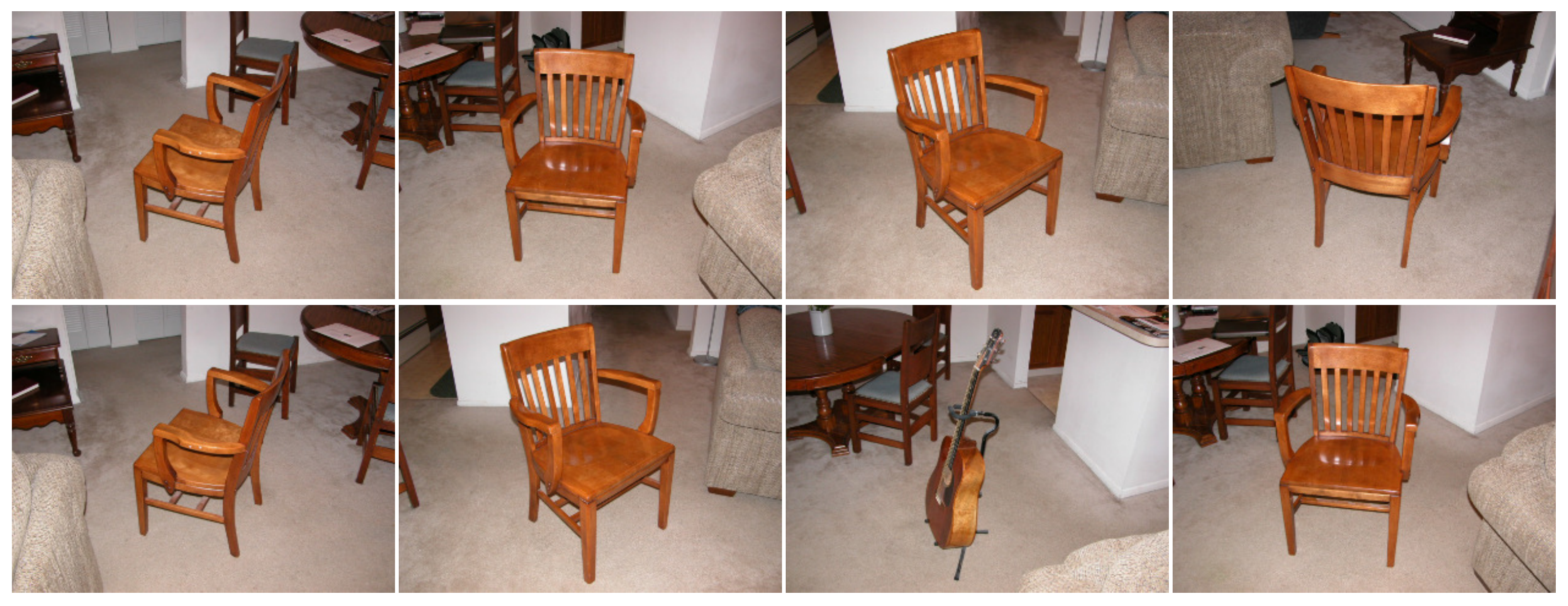}}
	\subfigure[\label{fig:ill:8}]{\includegraphics[width=0.40\textwidth]{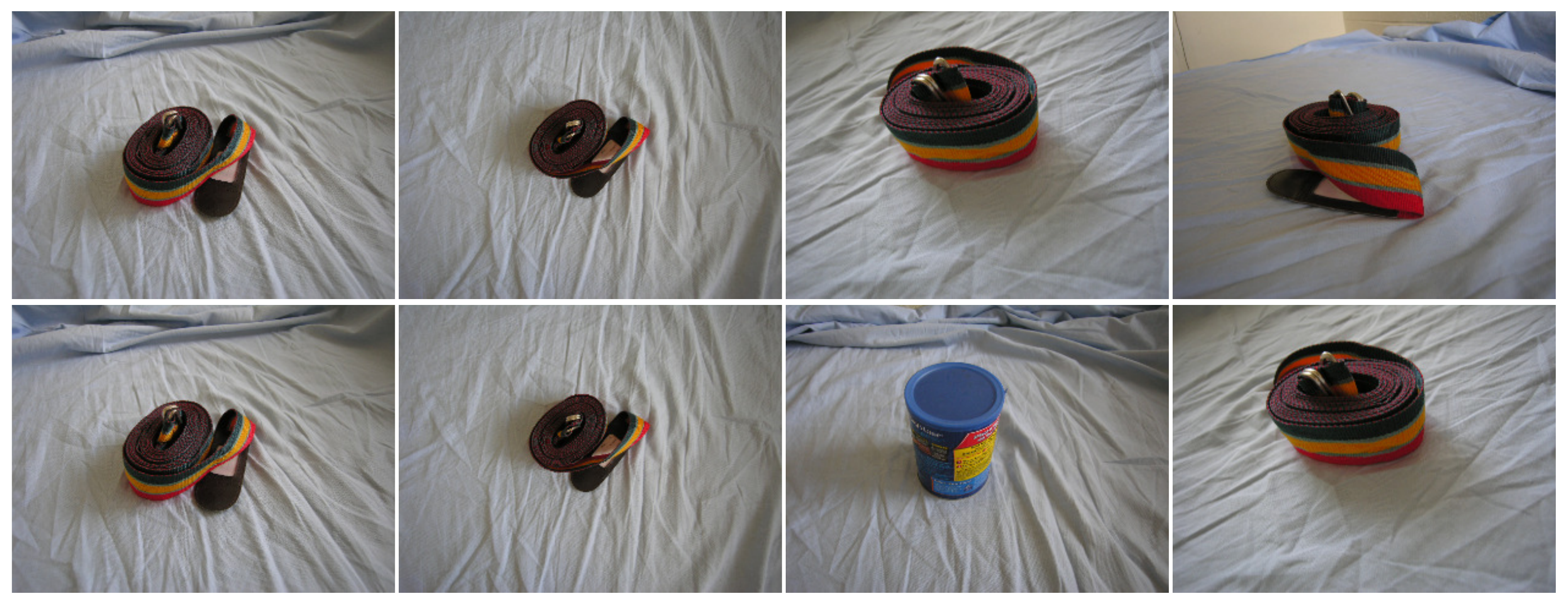}}\\
	\caption{Top 4 retrieval results on the UK-Bench dataset with
          the CNN and TICNN models. In each group, the results of the
          CNN model and TICNN model are given in the top and bottom
          row, respectively. The results in (a)-(f) are positive cases, whereas (g)-(h) show two failure cases.}
	\label{fig:ill}
\end{figure*}

We also evaluate our TICNN model on the popular image retrieval
benchmark dataset UK-Bench \cite{uk-bench}. This dataset includes $2550$
groups of images, each containing $4$ relevant samples about one certain
object or scene from different viewpoints. Each of the total $10200$
images is used as one query to perform image retrieval targeting at
finding all its $3$ counterparts. We choose UK-Bench because the
viewpoint variation in the dataset is very common. Although many of
the variation types are beyond the three types of geometry
transformations that we attempt to address in this paper, we demonstrate the effectiveness of TICNN for solving many severe rotation, translation and scale variance cases in image retrieval tasks. To test the CNNs in a dataset with larger scale, we also add the $1M$ images in MIR Flickr \cite{huiskes08} as negative examples.

For feature extraction, we follow the practice proposed in
\cite{interactive} to feed each image into the TICNN and CNN models,
and we perform average pooling on the last pooling layer to obtain a
compact feature representation. Specifically, we apply the AlexNet
architecture for the baseline CNN model and the TICNN model, of which
the fifth pooling layer generates a $13\times13\times256$ shape
response map. We obtain the average values for each of the 256
channels and obtain a 256-dimensional feature for every
image. Subsequently, we compute the root value of each dimension and perform $L2$ normalization.

To perform image retrieval in UK-Bench, the Euclidean distances of the query image with respect to all $10200$ database images are computed and sorted.
Images with the smallest distances are returned as top ranked
images. The NS-Score (average top four accuracy) is used to evaluate
the performance, and a score of $4.0$ means that all the relevant images are successfully retrieved in the top four results.

In the evaluation, we compare our TI-Conv1,2,5 model with the AlexNet
model trained on ImageNet. As shown in Table
\ref{perf:ukbench-result}, TICNN achieves a considerable performance
improvement. After carefully analyzing the retrieved results, we find
that our model successfully solves multiple transformation cases in
the retrieval task. In Fig. \ref{fig:ill}, we illustrate six positive
cases in which TICNN outperforms the baseline in
Figs. \ref{fig:ill:1}-\ref{fig:ill:6}, and we also provide two negative cases in Fig. \ref{fig:ill:7}-\ref{fig:ill:8}.

\begin{table}[t]
\begin{center}
\begin{tabular}{|l|c|c|}
\hline
 Model & UK-Bench & UK-Bench+MIRFlickr \\
 \hline
TICNN (R) & \textbf{3.572} & \textbf{3.483}\\
\hline
TICNN (T) & 3.536 & 3.452 \\
\hline
TICNN (S) & 3.551 & 3.461 \\
\hline
TICNN (RTS) & 3.561 & 3.479\\
\hline
\hline
CNN & 3.518 & 3.350 \\
\hline
SIFT & 3.350 & 3.295\\
\hline
\end{tabular}
\end{center}
\caption{Performance of the TICNN and CNN models on the UK-Bench retrieval dataset. TICNN has the same architecture as AlexNet. R, T, S and RTS represent ration, translation, scale, and rotation-translation-scale, respectively. }
\label{perf:ukbench-result}
\end{table}

As shown in Fig. \ref{fig:ill:1}, the CNN model only retrieves one
relevant image, while TICNN retrieves another one as shown in the
bottom row. The additionally returned database image has an obvious
scale transformation with respect to the query, and it is successfully
retrieved as the most similar one in the database by our model. As
shown in Fig. \ref{fig:ill:2} and Fig. \ref{fig:ill:3}, our model also performs better for rotation variance than CNN. In Fig. \ref{fig:ill:4} and Fig. \ref{fig:ill:5}, the objects (the road sign and basketball) appear in different scales and different spatial locations, which impacts the performance of CNN but is well handled by TICNN.

Regarding the failure cases of TICNN shown in Fig. \ref{fig:ill:7},
our model returns a false positive sample as shown in the third column
of the bottom row. We observe that this false positive image also
contains a chair like the query, but it appears in different scales
and locations, which leads to a false match. Some other failure cases
are similar to that in Fig. \ref{fig:ill:8}, where a large 3D viewpoint variance occurs, which is impossible to be produced in our training phase.

\section{Conclusions}
\label{sec:conclusion}
In this paper, we introduce a very simple and effective approach to improve the transform invariance of CNN models. By randomly
transforming the feature maps of CNN layers during training, the
dependency of the specific transform level of the input is reduced.
Our architecture is different from that of previous approaches because we improve the invariance of deep learning models without
adding any extra feature extraction modules, any learnable parameters
or any transformations on the training dataset.
Therefore, the transform invariances of current CNN models are very easy to be improved by just replacing their corresponding
weights with our trained model. Experiments show that our model outperforms CNN models in both image recognition and
image retrieval tasks.


\textbf{Acknowledgments} This work is supported by NSFC under the contracts No.61572451 and No.61390514, the 973 project under the contract No.2015CB351803, the Youth Innovation Promotion Association CAS CX2100060016, Fok Ying Tung Education Foundation, Australian Research Council Projects: DP-140102164, FT-130101457, and LE140100061.
%
\bibliographystyle{abbrv}
\bibliography{ref}  
%
%

\end{document}